\documentclass[letterpaper]{article} % DO NOT CHANGE THIS
\usepackage{aaai2026}  % DO NOT CHANGE THIS
\usepackage{times}  % DO NOT CHANGE THIS
\usepackage{helvet}  % DO NOT CHANGE THIS
\usepackage{courier}  % DO NOT CHANGE THIS
\usepackage[hyphens]{url}  % DO NOT CHANGE THIS
\usepackage{graphicx} % DO NOT CHANGE THIS
\usepackage{natbib}  % DO NOT CHANGE THIS AND DO NOT ADD ANY OPTIONS TO IT
\usepackage{caption} % DO NOT CHANGE THIS AND DO NOT ADD ANY OPTIONS TO IT
\usepackage{algorithm}
\usepackage{algorithmic}

\usepackage{newfloat}
\usepackage{listings}
\DeclareCaptionStyle{ruled}{labelfont=normalfont,labelsep=colon,strut=off} % DO NOT CHANGE THIS
\floatstyle{ruled}
\newfloat{listing}{tb}{lst}{}
\floatname{listing}{Listing}
\usepackage[table]{xcolor}

\title{AI Copilots for Reproducibility in Science: A Case Study}
\author{
    Adrien Bibal\textsuperscript{1},
    Steven N. Minton\textsuperscript{1},
    Deborah Khider\textsuperscript{2},
    Yolanda Gil\textsuperscript{2}
}
\affiliations{
    \textsuperscript{\rm 1} InferLink Corporation, Los Angeles, California, USA\\
    \textsuperscript{\rm 2} Information Sciences Institute (ISI), University of Southern California (USC), Los Angeles, California, USA\\
    abibal@inferlink.com, sminton@inferlink.com, khider@usc.edu, gil@isi.edu
}

\begin{document}

\maketitle

\begin{abstract}
Open science initiatives seek to make research outputs more transparent, accessible, and reusable, but ensuring that published findings can be independently reproduced remains a persistent challenge. In this paper we describe an AI-driven "Reproducibility Copilot" that analyzes manuscripts, code, and supplementary materials to generate structured Jupyter Notebooks and recommendations aimed at facilitating computational, or ``rote'', reproducibility. Our initial results suggest that the copilot has the potential to substantially reduce reproduction time (in one case from over 30 hours to about 1 hour) while achieving high coverage of figures, tables, and results suitable for computational reproduction. The system systematically detects barriers to reproducibility, including missing values for hyperparameters, undocumented preprocessing steps, and incomplete or inaccessible datasets. Although preliminary, these findings suggest that AI tools can meaningfully reduce the burden of reproducibility efforts and contribute to more transparent and verifiable scientific communication.
\end{abstract}

\begin{links}
    \link{Code}{https://bitbucket.org/inferlink/reproducibility_copilot}
    \link{Notebooks}{https://zenodo.org/records/17162894}
    % \link{Extended version}{https://aaai.org/example/extended-version}
\end{links}

\section*{Introduction}

Open science has emerged as a transformative movement aimed at improving the transparency, accessibility, and reliability of scientific knowledge~\cite{OpenScienceBook}. Its importance stems from the recognition that science advances most effectively when findings, methods, and data are openly shared and critically examined. Open science strengthens the scientific enterprise by making it easier for readers to transparently investigate published results and build on prior work, thereby accelerating discovery and enhancing the reliability of scientific claims.

Among the pillars of open science, reproducibility is particularly central. Reproducibility ensures that scientific results are not isolated artifacts but can be independently verified and built upon by others. Without it, trust in scientific claims is weakened and progress becomes difficult to assess or scale. Although some researchers have long championed open practices~\cite{boai2002,mckiernan2016open,lowndes2017our}, the past decade has seen a significant acceleration, driven in part by mandates from funders and publishers~\cite{nugroho2015comparison,vandeneyden2016survey,zuiderwijk2014open,doerr2019giving}, and by investments in open science infrastructure - such as curated data repositories, community standards, and tools such as electronic notebooks and containers - alongside the automation workflows that streamline the use of these tools by scientists.

As discussed by~\citet{GeosciencePaperOfTheFuture}, the practice of open science can require additional time and work by researchers, such as documenting code and data so that analyses can be reproduced. However, artificial intelligence (AI), as well as electronic notebooks and containers, can now help researchers in this endeavor. Indeed, the striking new advances in AI in recent years, including in large language models (LLMs), have created new opportunities to assist researchers and potentially revolutionize current paradigms, including reviewing, curating, publishing, distributing and sharing new technical results (see, for instance, the work of~\citet{Paper2Agent} and the new Agents4Science conference\footnote{\url{https://agents4science.stanford.edu/}}). To take advantage of these capabilities, we have been developing an AI-copilot framework called OpenPub to help authors and readers, but also reviewers and publishers, in the open science enterprise.

In this paper, we present a case study on an AI copilot that 1) helps authors make their research more reproducible and 2) helps readers to reproduce the authors' research. We developed an initial version of this {\em Reproducibility Copilot} and demonstrated the feasibility of the approach. For instance, we revisited a study on reproducibility by~\citet{gundersen2025unreasonableeffectiveness}, and showed that for a paper that previously took 30 hours to reproduce, our copilot could produce recommendations which, if followed, would reduce the time to reproduce the paper to approximately one hour.

\section*{Use Case: A Copilot for Reproducing Research}

The ``Reproducibility Copilot'' that we developed serves two complementary functions: assisting authors in making their work more reproducible and guiding readers through the process of reproducing the work. The prototype was designed to operate on typical research artifacts (manuscripts, supplementary material, code, and datasets) and generate customized guidance, such as a Jupyter Notebook for readers or annotated comments in PDFs for authors.

In considering what problems the Reproducibility Copilot should tackle, we were influenced by the analysis of~\citet{gundersen2025unreasonableeffectiveness}, who analyzed the impact of different issues related to making science more open. Their categorization included impactful issues, such as missing hyperparameter values for reproducing experiments, missing or undocumented code components, and incomplete data references. Using~\citet{gundersen2025unreasonableeffectiveness} analysis as a guide, we designed a copilot capable of detecting and providing recommendations to fix a spectrum of problems.

\begin{figure*}[t]
    \centering
    \includegraphics[width=0.8\textwidth]{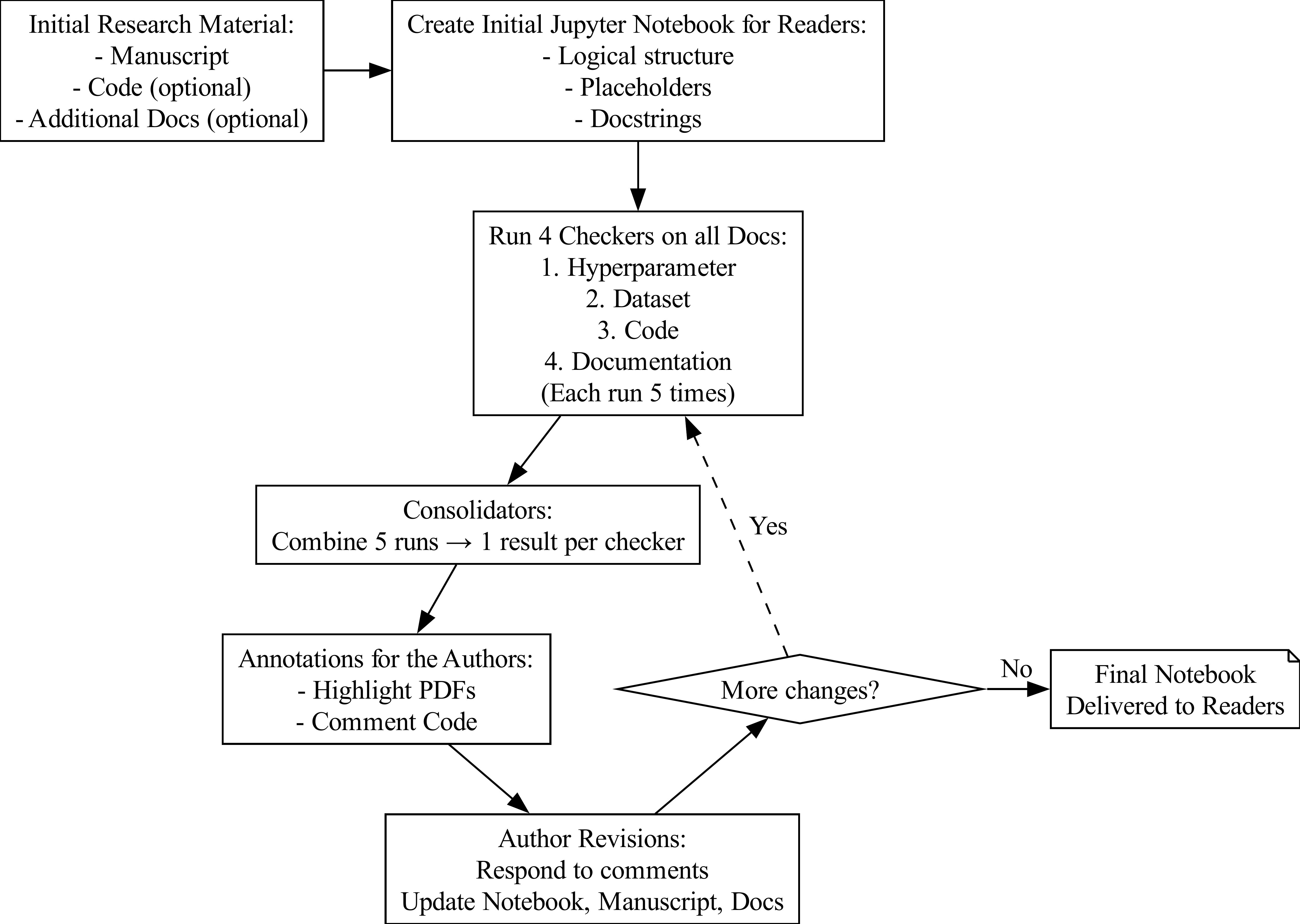}
    \caption{System architecture of our AI-driven Reproducibility Copilot, illustrating its dual-role design that simultaneously supports authors and readers. While code and additional docs enhance results, they are optional - the system can function with the manuscript alone, generating PDF highlights to identify missing reproducibility information.}
    \label{fig:architecture}
\end{figure*}

Figure~\ref{fig:architecture} provides a high-level overview of the architecture of the system, which is designed to support both authors and readers. This dual-role design allows the system to anticipate the reader’s perspective, identify elements that can be difficult to interpret or reproduce, and generate a targeted output for the reader (e.g., a Jupyter Notebook to guide reproduction) while also producing actionable recommendations for the author (e.g., comments in the manuscript and code annotations). We observed that the iterative process of our system typically converges within one or two iterations. This rapid stabilization occurs because the LLM handles the majority of the interpretative workload upfront, while simultaneously prompting the author to supply the bulk of any missing context during the initial interaction. Consequently, subsequent iterations yield diminishing marginal gains.

One of the first steps of the Reproducibility Copilot is to generate a reproducibility-focused Jupyter Notebook for the reader. This Notebook is structured around the logical flow of the original study, using placeholders where information is missing from the authors. These placeholders serve not only to guide the reader through incomplete parts of the research, but also to inform the author about where clarifications or additions are needed. Anchoring the LLM analysis on this Notebook helps focus the checks on the most essential elements for reproduction.

\begin{figure*}[t]
    \centering
    \includegraphics[width=0.8\textwidth]{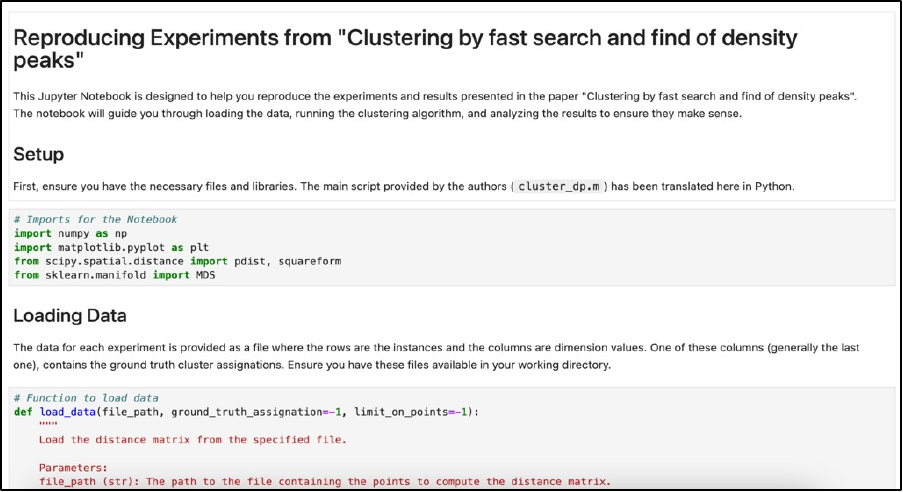}
    \caption{Opening section of the Jupyter Notebook automatically generated by the system to guide readers in reproducing the experiments from~\citet{rodriguez2014clustering}.}
    \label{fig:example_notebook}
\end{figure*}

\begin{figure*}[t]
    \centering
    \includegraphics[width=0.8\textwidth]{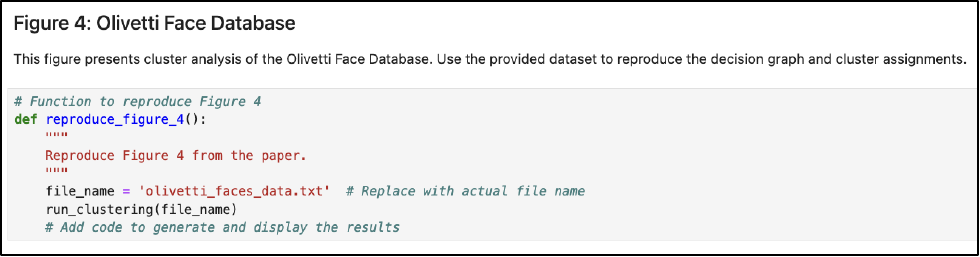}
    \caption{Excerpt from the system-generated Notebook for reproducing Figure 4 from~\citet{rodriguez2014clustering}. Since the original code for this figure was not provided, the system inserts minimal placeholder content to signal the missing implementation.}
    \label{fig:example_notebook_cell}
\end{figure*}

The generated Notebook, shown in Figure~\ref{fig:example_notebook}, focuses on the reproduction of the figures and tables of the study.\footnote{We have observed that most results tend to appear in figures or tables. So, for simplicity, results that are not mentioned in any table or figure are effectively ignored in the current implementation.} By organizing the reader’s process and clarifying where input is needed, the Notebook serves as a scaffold rather than a replacement for the author’s work. In Figure~\ref{fig:example_notebook_cell}, we show an example cell that corresponds to a figure from~\citet{rodriguez2014clustering}. The system has added a placeholder to indicate that the code to generate the plot is missing (see ``\# Add code to generate and display the results''), and uses this signal to later generate a recommendation to the author to supply the relevant code.

Once the Notebook is generated, the system applies a series of dedicated modules to identify and flag missing or ambiguous information. Through successive design iterations, four modules were designed and implemented in our prototype:
\begin{enumerate}
    \item Hyperparameter Checker: Analyzes code and documentation to identify critical hyperparameter values that are missing or ambiguous. Recommends clarifications to the authors.
    \item Dataset Checker: Scans the manuscript and code for data usage, verifying the presence of direct links to datasets. Flags missing datasets and proposes resolution steps.
    \item Code Checker: Examines the code to detect missing snippets essential for experiment replication.
    \item Documentation Checker: Assesses inline comments and code structure to evaluate readability and comprehensibility. Generates clarifications for unclear segments.
\end{enumerate}

Each of these modules was implemented using GPT-4o~\cite{GPT-4o} and a 2-step prompt strategy. For the first step, a prompt was designed to list candidates in each specific module. For instance, in the case of the Hyperparameter Checker, this first step consisted in listing all the hyperparameters mentioned in the manuscript and code. The second step then uses a specific prompt to reduce the list according to the issue at hand and provides some recommendations to the authors. In the case of the Hyperparameter Checker, this consisted in keeping the hyperparameters for which some values are not clearly provided for some experiments. Given the non-deterministic nature of GPT-4o, these checks were performed five times, and another prompt was used with GPT-4o to consolidate these results by taking the union of the final lists produced.

\begin{figure*}[t]
    \centering
    \includegraphics[width=0.8\linewidth]{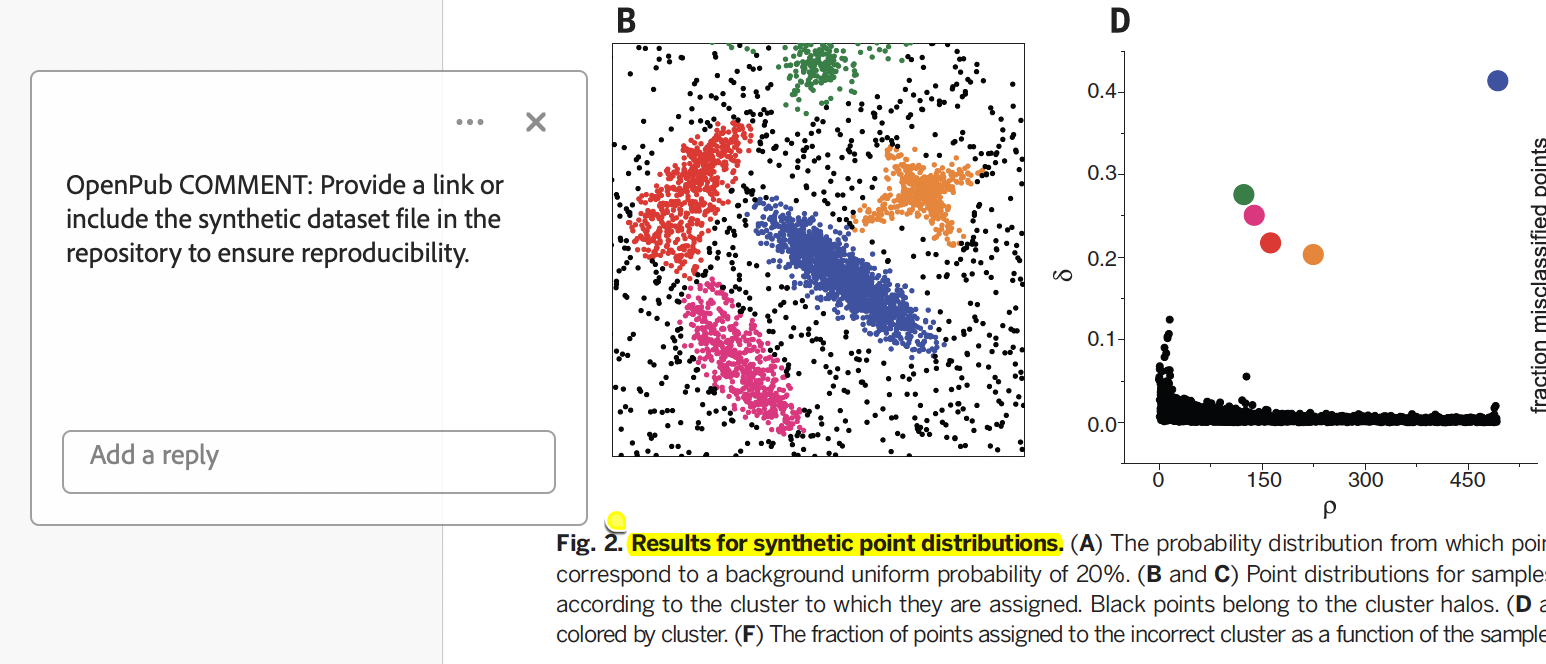}
    \caption{Example of a highlight produced by the copilot on the manuscript PDF of~\citet{rodriguez2014clustering}. The associated comment suggests including data to ensure the reproducibility of the authors' results.}
    \label{fig:example_comment}
\end{figure*}

Once all module outputs are consolidated, the final list produced by each module is used by the system to annotate the manuscript PDF with targeted highlights and margin comments (see, e.g., Figure~\ref{fig:example_comment}), and inserts inline feedback within code files. As the number of comments can be high, depending on the reproducibility issues in the paper, the corresponding feedback comments were deliberately kept minimal to avoid overwhelming the authors.

\section*{Results}

A crucial result of our case study is the demonstration of a significant time reduction in reproducing scientific research. To demonstrate the practical impact of the system, we selected a paper by~\citet{rodriguez2014clustering}, which had previously been analyzed in detail by~\citet{gundersen2025unreasonableeffectiveness}. This paper, which introduced a new machine learning method for clustering, was chosen specifically because it was one of the studies successfully reproduced by Gundersen et al., confirming that replication of the results was possible. Furthermore, the reported reproduction time of 33 hours by Gundersen et al. was below the 40-hour cutoff used in their study. Since this reproduction effort did not reach the cutoff limit, the 33-hour figure reflects the complete time it took to fully reproduce the study, making it a reliable benchmark against which we could measure our system's efficiency. 

Although it took a researcher 33 hours to reproduce the experiments in the Gundersen et al. study, using our copilot system we found that users could complete the task in a relatively short time. Specifically, we asked two individuals (who were unfamiliar with the reproduced paper, its methodology, and its data) to independently play the role of readers and use our copilot to replicate the study. The first user, a researcher with a Ph.D. in data science, completed the task in about 1 hour. The second, a somewhat less experienced user (a college student in computer science with substantial AI experience), also achieved similar performance (i.e., about 1 hour to reproduce the experiments), suggesting that the system makes reproducibility more accessible, not just more efficient.

These tests were carried out under the assumption that the author is fully compliant with the recommendations generated by our tool, allowing us to evaluate the effectiveness of the system in an ideal usage scenario. What this means is that the copilot generated suggestions for how to modify the code and documentation, then we (members of our project team) played the role of the authors, modifying and/or adding code, data, and documentation according to the system’s suggestions. To emulate the role of the authors, we reused and adapted publicly available code: the supplementary material provided by~\citet{rodriguez2014clustering} and the implementation from~\citet{gundersen2025unreasonableeffectiveness}~\footnote{available at \url{https://github.com/AIReproducibility2018/ClusteringByFastSearchAndFindOfDensityPeaks}}. Once the copilot was satisfied, the paper and supplementary material were made available to the user we recruited for our reproducibility task. In practice, for the paper by~\citet{rodriguez2014clustering}, working with our system produced the data for each experiment and a Notebook containing functions to load each file in the right format, specific hyperparameter values, and usage explanations. The system facilitated this by highlighting relevant sections of the original code and manuscript, prompting the author to clarify ambiguities and populate the Notebook with the correct context. Because the author was prompted to address these details, the recruited users were spared the manual reverse-engineering that made the reproduction of the paper take more than 30 hours in the study by~\citet{gundersen2025unreasonableeffectiveness}.

As a second feasibility test, we evaluated the coverage of the system’s content, which means its ability to identify and scaffold the core experimental content required for reproduction. To operationalize this, we focused on results presented in tables and figures, as these are clearer targets than smaller claims embedded in the manuscript text. This focus allowed us to systematically compare how much reproducible content the system is able to detect and guide users through.

To evaluate content coverage, we conducted a separate experiment in which we compared the Jupyter Notebook generated by our copilot with one that was manually created. This experiment focused on an Earth Sciences study, in part to test our prototype on another type of subject. Specifically, the study used Jupyter Notebooks that were developed by~\citet{Khider_Coral_Sr_Ca_Calibration_2024} (one of the coauthors of this paper) to help teach the research behind a study by~\citet{Tortugas}. The goal of these pedagogical Notebooks was not full reproducibility, but rather a guide for early-career scientists to learn to use data science libraries such as scikit-learn to perform a calibration task. While the original study focused on the robustness of the proxy signal in space and among different species of corals, the pedagogical Notebooks illustrate the use of Frequentist and Bayesian approaches in performing this calibration task. In this case, to mimic the original authors following the system's suggestions, we adapted the code of Dr. Deborah Khider~\footnote{available at \url{https://khider.github.io/dry-tortugas-calibration-fun}}.

When comparing the Notebook produced by our copilot with the pedagogical ones from Dr. Khider, we found that our copilot provided a wide coverage of figures and tables, capturing not only the core figures and tables highlighted in the educational Notebooks, but also additional elements of the paper that are essential for full reproduction. As shown in Table~\ref{tab:coverage_results}, our copilot covered all key figures and tables included in the educational Notebooks (green rows), as well as several additional visualizations and results that are necessary from a reproducibility point of view, but were excluded in the pedagogical version (white rows). The only omission from the coverage of our copilot was a figure (Figure 8 in~\citet{Tortugas}), which was also not prioritized in the teaching-oriented Notebooks (orange row). This comprehensive coverage demonstrates that such an AI system does not merely replicate what educators consider useful for instruction: it successfully identifies and scaffolds the full scope of reproducible content in a research paper, showing that this kind of tool can be quite complete when used for scientific validation.

\begin{table*}[!ht]
    \centering
    \begin{tabular}{c|c|c}
        \shortstack{Figures/Tables\\ from~\citet{Tortugas}} & Our Copilot’s Notebook & Educational Notebooks\\\hline
        Table 2 & $\surd$ & $\times$\\
        Figure 1 & $\surd$ & $\times$\\
        \rowcolor{green!20}
        Figure 2 & $\surd$ & $\surd$\\
        Figure 3 & $\surd$ & $\times$\\
        Figure 4 & $\surd$ & $\times$\\
        \rowcolor{green!20}
        Figure 5 & $\surd$ & $\surd$\\
        Figure 6 & $\surd$ & $\times$\\
        \rowcolor{green!20}
        Figure 7 & $\surd$ & $\surd$\\
        \rowcolor{orange!30}
        Figure 8 & $\times$ & $\times$\\
        Figure 9 & $\surd$ & $\times$\\
        \rowcolor{green!20}
        Table 3 & $\surd$ & $\surd$
    \end{tabular}
    \caption{Comparative analysis of reproducibility coverage between the Notebook generated by our copilot and Dr. Khider’s educational Notebooks~\cite{Khider_Coral_Sr_Ca_Calibration_2024} for the~\citet{Tortugas} study. Green rows indicate figures and tables covered by both systems, orange rows mark a figure not covered by our copilot but also excluded from the educational Notebooks, and white rows highlight additional figures and tables captured by our copilot but omitted from the educational Notebooks.}
    \label{tab:coverage_results}
\end{table*}

\section*{Discussion}

During our research, we identified several future challenges related to the use of AI to enhance reproducibility.

\paragraph{Scientific Understanding} Reproducibility can take many forms, such as computational reproducibility~\cite{freire2012computational,piccolo2016tools,gruning2018practical}, empirical reproducibility~\cite{stodden2013resolving}, and outcome, analysis and interpretation reproducibility~\cite{gundersen2022sourcesOfirreproducibility}. Throughout the use case in this paper, we focused on what can be called ``rote reproducibility'', which we define as regenerating figures and tables using the same code and data. This type of reproducibility is only a small part of full ``scientific reproducibility'', where the reconstruction of the reasoning and conclusions behind the research is sought. However, achieving complete scientific reproducibility is substantially more difficult as it requires access to the logic, rationale, and underlying assumptions that may not be explicitly stated in a paper. 

In this case study, only preliminary steps were taken towards the challenging issues of scientific reproducibility. For example, we inferred analytical workflows from the manuscript and added preliminary instructional steps in the generated Notebook to guide readers in evaluating the plausibility of reproduced results. Developing a copilot with a full understanding of the science behind the paper is a challenge to tackle in future work and would require knowledge beyond analysis.

\paragraph{User Adaptation} A related complication comes from differences in the scientific background of the readers. What may be intuitive to an expert may be opaque to a student or newcomer. The tailoring of the guide based on the reader’s expertise is essential to promote scientific understanding. In future work, we plan to automatically construct semantic representations of the author analyses, enabling AI copilots to generate reader-specific guidance. This would not only support reproducibility, but also accelerate comprehension and learning for reviewers and readers alike. Indeed, our observations during our study suggest that simply understanding a paper often requires as much effort as reproducing the study using the Notebook produced by our copilot.

\paragraph{Data Preparation and Diagnostic Analysis} Another prominent challenge is related to data formatting and preprocessing. When input data are not properly structured to work with the authors code, or worse, if essential preprocessing steps are missing, it becomes time-consuming for readers to identify and resolve these issues. At the same time, it remains difficult for an LLM to infer from raw numeric files that something is amiss. For instance, a file might appear valid on the surface, yet fail silently due to missing normalization. There are a number of ways to address this problem. For instance, if a paper explicitly states certain requirements on the input data (e.g., the data should have zero mean and unit variance), the system could test for these conditions and flag discrepancies. The system could also generate diagnostic code to validate whether the data meet the requirements specified by the authors. A related challenge is verifying that the output is reasonable/correct, i.e., analyzing the executed cells in the generated Notebook. While it may not always be feasible to verify whether generated results exactly match those in the original paper, at a minimum the system should at least detect major inconsistencies. For example, if the paper claims that five clusters were found in a dataset, but the system's reproduction produces none, that discrepancy could prompt the system to recommend further investigation.

\paragraph{Claim Identification} A further challenge lies in identifying which claims in a manuscript warrant reproduction. While some are clearly marked in tables and figures - such as reported accuracy scores or performance benchmarks - others are embedded subtly in narrative text. Furthermore, different types of claim can coexist in the paper. Claims can be stated at the end of the paper and made based on the results of the study, or at the beginning to often be contrasted with each other and show where the new study adds knowledge. These latter types of claim often lead to the hypothesis to be tested in the paper. Finally, there are assumptions that are often made in scientific studies that could be presented as claims (e.g., ``following the study by X et al., we will use the following method that has been proven valid for our data''). Building the capability to pinpoint all the claims, identify the ones that are reproducible, and verify that the supporting data and code are present is challenging.

\paragraph{Code Generation} Reproducing a study often (but not always) involves executing the code provided by the author. In some cases, there are bugs in that code, and in many cases, some parts of the code are missing, or at least, there is code that the reader needs to write to conveniently reproduce the work. Though in many cases we found that LLMs could do a fairly good job of automatically generating such code, we felt that this would be a distraction in this case study (especially if the generated code contained subtle errors), and we decided to rely on the authors to enter the code in the Notebook (who often actually has the code somewhere). In future work, we plan to reconsider how our copilot can make use of the rapidly improving capabilities of modern LLMs to help with bug-free code generation as part of the reproducibility process.

\paragraph{Multi-Turn Interactions} Similar to coding agents like Codex of OpenAI~\cite{Codex} and Claude Code of Anthropic~\cite{Claude_Code}, our Reproducibility Copilot could be enhanced with multi-turn interactions. In its current form, the copilot identifies potential improvements, allowing authors to revise their material for re-evaluation. However, challenges may arise if authors misinterpret the copilot's recommendations or require specific guidance on implementation (e.g., how to properly archive data). To address this, the system would benefit from a conversational interface that assists authors in resolving these reproducibility barriers interactively.

\section*{Limitations of Evaluation}

This study presents a feasibility study rather than a large-scale evaluation. A comprehensive evaluation could involve a controlled study where multiple PhD students reproduce papers in their respective fields, comparing the performance of those using the copilot against a control group working manually. Furthermore, our experiments assumed full compliance by the original authors with the system, which may not always be realistic in practice. Finally, the use of large language models introduces some variability due to non-deterministic outputs. These limitations underscore the preliminary nature of our findings, but also highlight clear directions for future work.

\section*{Conclusion}

This use case has demonstrated the feasibility and potential of AI-driven copilot systems to meaningfully improve the openness of scientific research. We demonstrated the impact of such an AI-based approach, as well as its capacity to deliver actionable impact across different user roles, in particular authors and readers. As a concrete example, we focused this initial effort on reproducibility, an area of open science that remains difficult, yet critical to address.

As shown in the study by~\citet{gundersen2025unreasonableeffectiveness}, current barriers to reproducibility often stem from missing or unclear information. In our evaluation, we showed that in a case documented in Gundersen et al., our AI system reduced the reproduction time from more than 30 hours to approximately 1 hour. This efficiency gain illustrates the value of AI-generated proactive guidance tailored to both authors and readers.

Importantly, the success of the system stems from its ability to surface specific, high-impact shortcomings - missing values for hyperparameters, inaccessible datasets, ambiguous code - and to communicate those deficiencies in ways that are actionable for authors. It also provides reproducibility scaffolds for readers, such as structured Jupyter Notebooks.

The broader implication of our findings is that reproducibility, often seen as a post-publication burden, can instead be integrated as a dynamic part of the scientific workflow. By helping authors anticipate reader needs and giving readers tools to navigate complex research materials, AI copilot systems can address complex problems by treating authors and readers as two faces of the same coin.

Although the system we developed for this use case focuses on ``rote reproducibility'' - the ability to regenerate results from the same datasets used in a study - future work should include pathways to support deeper, scientific reproducibility. These include modeling the logic and assumptions behind claims, adapting feedback to reader expertise, and even suggesting clarifications when results deviate.

\section*{Acknowledgments}
This material is based upon work supported by the Office of Naval Research under Contract No. N68335-25-C-0016. Any opinions, findings and conclusions or recommendations expressed in this material are those of the author(s) and do not necessarily reflect the views of the Office of Naval Research. 

\bibliography{biblio}

@book{OpenScienceBook,
  title={Open science: One term, five schools of thought},
  author={Fecher, Benedikt and Friesike, Sascha},
  booktitle={Opening Science},
  editors={Fecher, Benedikt and Friesike, Sascha},
  year={2014},
  publisher={Springer International Publishing}
}

@article{GeosciencePaperOfTheFuture,
  title={Toward the Geoscience Paper of the Future: Best practices for documenting and sharing research from data to software to provenance},
  author={Gil, Yolanda and David, C{\'e}dric H and Demir, Ibrahim and Essawy, Bakinam T and Fulweiler, Robinson W and Goodall, Jonathan L and Karlstrom, Leif and Lee, Huikyo and Mills, Heath J and Oh, Ji-Hyun and Pierce, Suzanne A and Pope, Allen and Tzeng, Mimi W and Villamizar, Sandra R and Yu, Xuan},
  journal={Earth and Space Science},
  volume={3},
  number={10},
  pages={388--415},
  year={2016}
}

@article{gundersen2025unreasonableeffectiveness,
  title={The unreasonable effectiveness of open science in AI: A replication study},
  author={Gundersen, Odd Erik and Cappelen, Odd and M{\o}ln{\aa}, Martin and Nilsen, Nicklas Grimstad},
  journal={Proceedings of the AAAI Conference on Artificial Intelligence},
  volume={39},
  number={25},
  pages={26211--26219},
  year={2025}
}

@misc{GPT-4o,
    author = "OpenAI",
    title = "Hello GPT-4o",
    url  = "https://openai.com/index/hello-gpt-4o/",
    year= "2024",
    addendum = "Accessed June 2nd, 2025"
}

@article{rodriguez2014clustering,
  title={Clustering by fast search and find of density peaks},
  author={Rodriguez, Alex and Laio, Alessandro},
  journal={Science},
  volume={344},
  number={6191},
  pages={1492--1496},
  year={2014}
}

@misc{Khider_Coral_Sr_Ca_Calibration_2024,
    author = {Khider, Deborah},
    title = {Coral Sr/Ca Calibration - An example from Dry Tortugas},
    url = {https://khider.github.io/dry-tortugas-calibration-fun/intro.html},
    version = {v1.1.0},
    year = {2024}
}

@article{Tortugas,
  title={A coral Sr/Ca calibration and replication study of two massive corals from the Gulf of Mexico},
  author={DeLong, Kristine L and Flannery, Jennifer A and Maupin, Christopher R and Poore, Richard Z and Quinn, Terrence M},
  journal={Palaeogeography, Palaeoclimatology, Palaeoecology},
  volume={307},
  number={1-4},
  pages={117--128},
  year={2011}
}

@article{gundersen2022sourcesOfirreproducibility,
  title={Sources of irreproducibility in machine learning: A review},
  author={Gundersen, Odd Erik and Coakley, Kevin and Kirkpatrick, Christine and Gil, Yolanda},
  journal={arXiv:2204.07610},
  year={2022}
}

@inproceedings{freire2012computational,
  title={Computational reproducibility: State-of-the-art, challenges, and database research opportunities},
  author={Freire, Juliana and Bonnet, Philippe and Shasha, Dennis},
  booktitle={Proceedings of ACM SIGMOD International Conference on Management of Data},
  pages={593--596},
  year={2012}
}

@article{piccolo2016tools,
  title={Tools and techniques for computational reproducibility},
  author={Piccolo, Stephen R and Frampton, Michael B},
  journal={Gigascience},
  volume={5},
  number={1},
  pages={s13742--016},
  year={2016}
}

@article{gruning2018practical,
  title={Practical computational reproducibility in the life sciences},
  author={Gr{\"u}ning, Bj{\"o}rn and Chilton, John and K{\"o}ster, Johannes and Dale, Ryan and Soranzo, Nicola and Van Den Beek, Marius and Goecks, Jeremy and Backofen, Rolf and Nekrutenko, Anton and Taylor, James},
  journal={Cell Systems},
  volume={6},
  number={6},
  pages={631--635},
  year={2018}
}

@article{stodden2013resolving,
  title={Resolving irreproducibility in empirical and computational research},
  author={Stodden, Victoria},
  journal={IMS Bulletin},
  volume={14},
  number={4},
  year={2013}
}

@article{nugroho2015comparison,
  title={A comparison of national open data policies: Lessons learned},
  author={Nugroho, Rininta Putri and Zuiderwijk, Anneke and Janssen, Marijn and de Jong, Martin},
  journal={Transforming government: People, process and policy},
  volume={9},
  number={3},
  pages={286--308},
  year={2015}
}

@misc{vandeneyden2016survey,
  title={Survey of Wellcome researchers and their attitudes to open research},
  author={Van den Eynden, Veerle and Knight, Gareth and Vlad, Anca and Radler, Barry and Tenopir, Carol and Leon, David and Manista, Frank and Whitworth, Jimmy and Corti, Louise},
  year={2016}
}

@article{zuiderwijk2014open,
  title={Open data policies, their implementation and impact: A framework for comparison},
  author={Zuiderwijk, Anneke and Janssen, Marijn},
  journal={Government information quarterly},
  volume={31},
  number={1},
  pages={17--29},
  year={2014}
}

@article{doerr2019giving,
  title={Giving software its due},
  author={Doerr, A and Rusk, N and Vogt, N and others},
  journal={Nature Methods},
  volume={16},
  pages={207},
  year={2019}
}

@article{lowndes2017our,
  title={Our path to better science in less time using open data science tools},
  author={Lowndes, Julia S Stewart and Best, Benjamin D and Scarborough, Courtney and Afflerbach, Jamie C and Frazier, Melanie R and O’hara, Casey C and Jiang, Ning and Halpern, Benjamin S},
  journal={Nature ecology \& evolution},
  volume={1},
  number={6},
  pages={0160},
  year={2017}
}

@article{mckiernan2016open,
  title={How open science helps researchers succeed},
  author={McKiernan, Erin C and Bourne, Philip E and Brown, C Titus and Buck, Stuart and Kenall, Amye and Lin, Jennifer and McDougall, Damon and Nosek, Brian A and Ram, Karthik and Soderberg, Courtney K and others},
  journal={elife},
  volume={5},
  pages={e16800},
  year={2016}
}

@misc{boai2002,
  title = {Budapest Open Access Initiative},
  author = {{Chan, Leslie} and {Cuplinskas, Darius} and {Eisen, Michael} and {Friend, Fred} and {Genova, Yana} and {Guédon, Jean-Claude} and {Hagemann, Melissa} and {Harnad, Stevan} and {Johnson, Rick} and {Kupryte, Rima} and {La Manna, Manfredi} and {Rév, István} and {Segbert, Monika} and {de Souza, Sidnei} and {Suber, Peter} and {Velterop, Jan}},
  year = {2002},
  howpublished = {\url{https://www.budapestopenaccessinitiative.org/read/}},
  note = {Accessed 2025-10-27}
}

@article{Paper2Agent,
  title={{Paper2Agent}: Reimagining research papers as interactive and reliable {AI} agents},
  author={Miao, Jiacheng and Davis, Joe R and Zhang, Yaohui and Pritchard, Jonathan K and Zou, James},
  journal={arXiv:2509.06917},
  year={2025}
}

@article{Codex,
  author       = {Mark Chen and
                  Jerry Tworek and
                  Heewoo Jun and
                  Qiming Yuan and
                  Henrique Pond{\'{e}} de Oliveira Pinto and
                  Jared Kaplan and
                  Harri Edwards and
                  Yuri Burda and
                  Nicholas Joseph and
                  Greg Brockman and
                  Alex Ray and
                  Raul Puri and
                  Gretchen Krueger and
                  Michael Petrov and
                  Heidy Khlaaf and
                  Girish Sastry and
                  Pamela Mishkin and
                  Brooke Chan and
                  Scott Gray and
                  Nick Ryder and
                  Mikhail Pavlov and
                  Alethea Power and
                  Lukasz Kaiser and
                  Mohammad Bavarian and
                  Clemens Winter and
                  Philippe Tillet and
                  Felipe Petroski Such and
                  Dave Cummings and
                  Matthias Plappert and
                  Fotios Chantzis and
                  Elizabeth Barnes and
                  Ariel Herbert{-}Voss and
                  William Hebgen Guss and
                  Alex Nichol and
                  Alex Paino and
                  Nikolas Tezak and
                  Jie Tang and
                  Igor Babuschkin and
                  Suchir Balaji and
                  Shantanu Jain and
                  William Saunders and
                  Christopher Hesse and
                  Andrew N. Carr and
                  Jan Leike and
                  Joshua Achiam and
                  Vedant Misra and
                  Evan Morikawa and
                  Alec Radford and
                  Matthew Knight and
                  Miles Brundage and
                  Mira Murati and
                  Katie Mayer and
                  Peter Welinder and
                  Bob McGrew and
                  Dario Amodei and
                  Sam McCandlish and
                  Ilya Sutskever and
                  Wojciech Zaremba},
  title        = {Evaluating Large Language Models Trained on Code},
  journal      = {arXiv:2107.0337},
  year         = {2021},
}

@misc{Claude_Code,
  title = {Claude 3.7 Sonnet and Claude Code},
  author = {Anthropic},
  year = {2025},
  howpublished = {\url{https://www.anthropic.com/news/claude-3-7-sonnet}},
  note = {Accessed 2025-12-08}
}


\end{document}